\documentclass[runningheads]{llncs}

\usepackage{iftex}
\ifPDFTeX
\usepackage[T1]{fontenc}
\usepackage[utf8]{inputenc}
\else
\usepackage{fontspec}
\usepackage{xeCJK}
\IfFontExistsTF{Songti SC}{\setCJKmainfont[Script=Default]{Songti SC}\setCJKmonofont[Script=Default]{Songti SC}}{\setCJKmainfont[Script=Default]{STSong}\setCJKmonofont[Script=Default]{STSong}}
\fi
\usepackage{graphicx}
\usepackage{amsmath}
\usepackage{array}
\usepackage{booktabs}
\usepackage{tabularx}
\usepackage{multirow}
\usepackage{arydshln}
\usepackage{url}
\usepackage{float}
\usepackage[hidelinks]{hyperref}
\begin{document}

\title{Beyond Single Character: Evaluating MLLMs for Sentence-Level Oracle Bone Inscription Understanding}
\titlerunning{}

\author{Ziqi Li\inst{2,*}\and
Zijian Chen\inst{1,3,*,}\textsuperscript{$\dag$} \and
Tingzhu Chen\inst{1,}\textsuperscript{$\dag$}\and
Guangtao Zhai\inst{1,3}} 
\authorrunning{Li et al.}
%
\institute{Shanghai Jiao Tong University, Shanghai 200240, China \and
Nanjing University of Science and Technology, Nanjing 210094, China
\\ 
\and
Shanghai Artificial Intelligence Laboratory, Shanghai 200232, China\\
\textsuperscript{*}These authors contribute equally \quad
\textsuperscript{$\dag$}Corresponding authors \\
\email{lzq0915@njust.edu.cn}\\
\email{\{zijian.chen,tingzhuchen,zhaiguangtao\}@sjtu.edu.cn}\\
\url{https://github.com/OBI-Future/S-OBI}
}
\maketitle

\begin{abstract}
Existing AI-assisted oracle bone inscription (OBI) visual recognition and understanding studies mainly focus on character-level, ignoring the long-form textual coherence and contextual dependencies embedded in complete divination charges. 
Recently, the powerful visual perception capabilities of multimodal large language models (MLLMs) have opened new possibilities for OBI information processing.
In this work, we introduce \textbf{S-OBI}, a novel benchmark for evaluating MLLMs in \underline{\textbf{S}}entence-level \underline{\textbf{OBI}} understanding. 
Instead of using noisy and incomplete rubbings as the visual input, S-OBI synthesizes clear and standardized sentence-level OBI instances through glyph substitution and composition. 
According to 95 original rubbings with translations that have been identified, corrected, and verified by experts, we replace characters in the original rubbings with corresponding clean glyph samples sourced from existing OBI datasets while preserving the overall inscriptional structure and semantic organization. This mitigates the influence of low-level distortions and enables a more focused evaluation of sentence-level OBI understanding.
Based on this, we design semantic matching, semantic slot extraction, and contextual reasoning tasks and obtain 695 question-answer pairs.
Experiments reveal the inferiority of contemporary MLLMs on sentence-level OBI understanding. 
In particular, visual perception errors in unmasked regions propagate through the reasoning chain, leading to erroneous predictions for masked characters, which indicates that sentence-level OBI understanding in current models remains strongly dependent on character-level recognition.
Overall, S-OBI provides a diagnostic benchmark for evaluating whether MLLMs can move beyond isolated character recognition toward structured inscription-level understanding.

\end{abstract}

\section{Introduction}
Oracle bone inscriptions (OBIs) are among the earliest mature writing systems in ancient China. They were mainly carved on turtle plastrons and animal bones in the late Shang dynasty and provide important evidence for studying early Chinese language, political organization, religious practices, and social structure \cite{obisurvey}. 
Unlike ordinary text recognition tasks, an OBI record is not simply a collection of isolated characters. A complete divinatory inscription usually follows a relatively structured form and may contain components such as the date, diviner, divination marker, charge, object, action, and sometimes missing or implicit contextual information. Therefore, OBI understanding requires more than recognizing individual glyphs. It also requires the model to integrate glyph perception, character order, semantic roles, and divinatory context to interpret the complete inscription. 

Traditional OBI studies mainly rely on expert interpretation. Researchers usually examine published materials, rubbings, hand copies, photographs, dictionaries, and archaeological context to decipher characters and interpret inscriptions~\cite{keightley1978sources}. This expert-led process provides reliable and explainable results, but it is time-consuming, highly dependent on specialist knowledge, and difficult to scale. 
With the development of computer vision and deep learning, OBI processing has gradually shifted from manual consultation to data-driven image recognition \cite{li2025obiformer}. Existing datasets and methods~\cite{obc306,oraclemnist,hustobc} have supported tasks such as OBI character detection, classification, retrieval, rejoining, and decipherment, enabling models to learn the correspondence between single-character images and modern Chinese characters or character categories. 
More recently, multimodal large language models (MLLMs) have been introduced into OBI-related tasks \cite{obibench,zhang2026specializing,zhou2026acse}, making it possible to explore the relation between visual glyph features, semantic knowledge, and cross-modal reasoning.

\begin{figure}[!t]
\centering
\includegraphics[width=\textwidth]{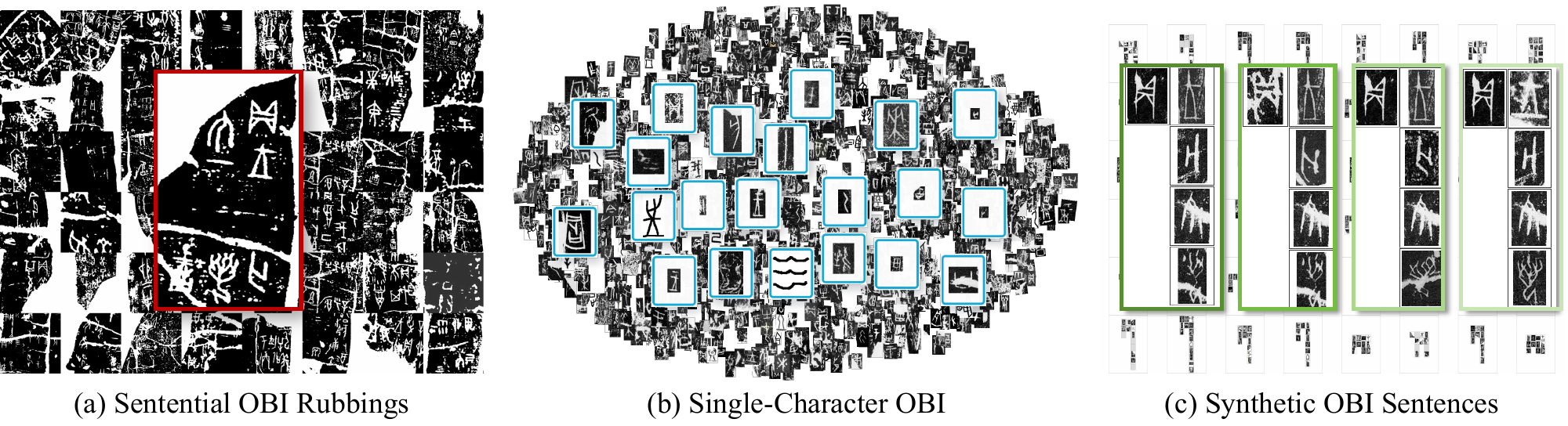}
\caption{Examples of OBI appearances covered in constructing the S-OBI. The red and green rectangles are paired original sentential OBI rubbings and the corresponding synthetic OBI sentences, respectively.}
\label{visualization}
\end{figure}

However, most existing studies focus on character-level or localized tasks. Their basic evaluation unit is typically a single glyph, with the primary objective being character identification or mapping the glyph to its modern Chinese equivalent. While these tasks are fundamental to OBI recognition and decipherment, they fall short of fully evaluating whether a model comprehends the internal structure of a complete divinatory inscription. For example, a model may recognize several individual characters but fail to identify their semantic roles in the inscription. It may also select a plausible interpretation without correctly recovering the reading order, punctuation boundaries, or missing characters. For a practical OBI understanding system, character recognition alone is not sufficient. The model should also be able to reason over sentence-level visual inputs and combine glyph identity, character order, divinatory structure, and contextual clues.

To address this limitation, we introduce S-OBI, a diagnostic benchmark for sentence-level OBI understanding. Unlike previous studies that mainly focus on isolated characters, S-OBI treats a complete divinatory inscription as the basic evaluation unit. It is built from 95 expert-corrected inscriptions and uses high-quality single-character OBI resources to reconstruct standardized sentence-level OBI images through glyph substitution. 
Fig. \ref{visualization} provides an example of the original sentential OBI rubbings, the single-character OBI pool, and the synthetic OBI sentences. 
This design reduces the influence of low-level visual noise and allows a more focused evaluation of sentence-level understanding. S-OBI contains 695 question-answer instances across three tasks, i.e., semantic matching, semantic slot extraction, and contextual reasoning. These tasks evaluate whether MLLMs can understand the overall meaning of an inscription, recover its structured semantic frame, and reason about punctuation boundaries and masked glyphs from context.

We systematically evaluate representative MLLMs on S-OBI. The results show that current models are still far from reliable sentence-level OBI understanding. Although some models can handle coarse semantic selection or reading-order choices to a certain extent, they struggle with structured semantic recovery, punctuation restoration, and masked glyph completion. These findings suggest that current MLLMs have not yet developed a robust ability to model the structure and context of complete divinatory inscriptions.

The main contributions of this paper are summarized as follows:
\begin{itemize}
    \item We propose a new sentence-level OBI understanding task. Different from previous studies that mainly focus on isolated characters, our task treats a complete divinatory inscription as the basic evaluation unit and requires models to jointly consider glyph perception, character order, semantic roles, and divinatory context.
    \item We construct S-OBI, a diagnostic benchmark for sentence-level OBI understanding. S-OBI reconstructs standardized sentence-level OBI images through glyph substitution using high-quality single-character OBI resources and provides 695 question-answer instances covering semantic matching, semantic slot extraction, and contextual reasoning.
    \item We evaluate 10 MLLMs on sentence-level OBI understanding. The results show that existing models can sometimes handle coarse semantic selection but still struggle with structured semantic recovery, punctuation restoration, and masked glyph completion, indicating that sentence-level OBI understanding remains a challenging problem for current MLLMs.
\end{itemize}

\section{Related Work}

\subsection{Oracle Bone Inscription Datasets}
Existing OBI datasets have primarily been developed for character-level recognition \cite{obibench}, classification \cite{li2026roots}, retrieval, generation \cite{li2025mitigating}, and decipherment tasks \cite{obisurvey}. 
Representative datasets, such as OBC306~\cite{obc306}, Oracle-MNIST~\cite{oraclemnist}, and HUST-OBC~\cite{hustobc}, contain large-scale oracle character images with category labels for recognition or decipherment, serving as a common basis for single-glyph evaluation. 
Another important line of character-level datasets usually pairs cropped OBI glyphs with their modern Chinese counterparts, textual glosses, or evolutionary forms, thereby providing supervised signals for bridging the historical gap between ancient and modern scripts. 
For example, EVOBC~\cite{evobc} models the evolutionary trajectory from oracle bone script to later Chinese scripts. OracleSem~\cite{oraclesage} enriches character-level samples with semantic annotations, including pictographic composition, structural organization, and semantic evolution. PictOBI-20k~\cite{pictobi} links oracle glyphs with real-world object images and question-answer pairs to evaluate MLLMs on visual decipherment of pictographic OBIs.

However, these evaluation artifacts remain predominantly isolated in character, which mainly answer questions, such as {\it Which character is this?} or {\it Which modern Chinese character may this glyph correspond to?}, while paying limited attention to inscription-level structure, character order, semantic roles, and contextual reasoning.

\subsection{OBI Understanding}
Early OBI understanding was primarily expert-led, with researchers manually examining published materials, rubbings, tracings, photographs, and ancient books to decipher characters and interpret inscriptions. This process laid the foundation for OBI studies, but it was time-consuming, highly dependent on specialist knowledge, and difficult to reproduce at scale \cite{obisurvey,li2026comprehensive}.
Recently, the rise of deep learning shifted OBI processing from expert-centric manual procedures to data-driven automatic frameworks. 
Deep neural networks have been widely applied to OBI character recognition, rejoining, classification, and retrieval, and Transformer-based architectures further improved the modeling of structural dependencies in degraded or fragmented glyphs \cite{liu2020oracle,yue2022dynamic,fujikawa2023recognition}. 
However, these methods are still mainly driven by visual features and often lack explicit modeling of the textual and semantic content of divinatory inscriptions. 
More recently, large multimodal models have introduced a new paradigm for cross-modal OBI perception and interpretation \cite{chen2025just}, enabling the field to move beyond isolated character recognition toward contextual decipherment. 
Representative efforts include OBI-Bench \cite{obibench}, OracleSage \cite{oraclesage}, OracleFusion \cite{li2025oraclefusion}, and related LMM-based OBI decipherment systems \cite{liu2026alphaoracle,li2025oracleagent}. Nevertheless, current LMMs remain fragile on noisy, original, and cross-modal OBI inputs, and their ability to perform structured sentence-level reasoning over complete divinatory inscriptions is still insufficient.
Our proposed S-OBI is designed to address this gap by treating the sentence-level OBI image sequences as the basic evaluation unit.

\begin{figure}[!t]
\centering
\includegraphics[width=\textwidth]{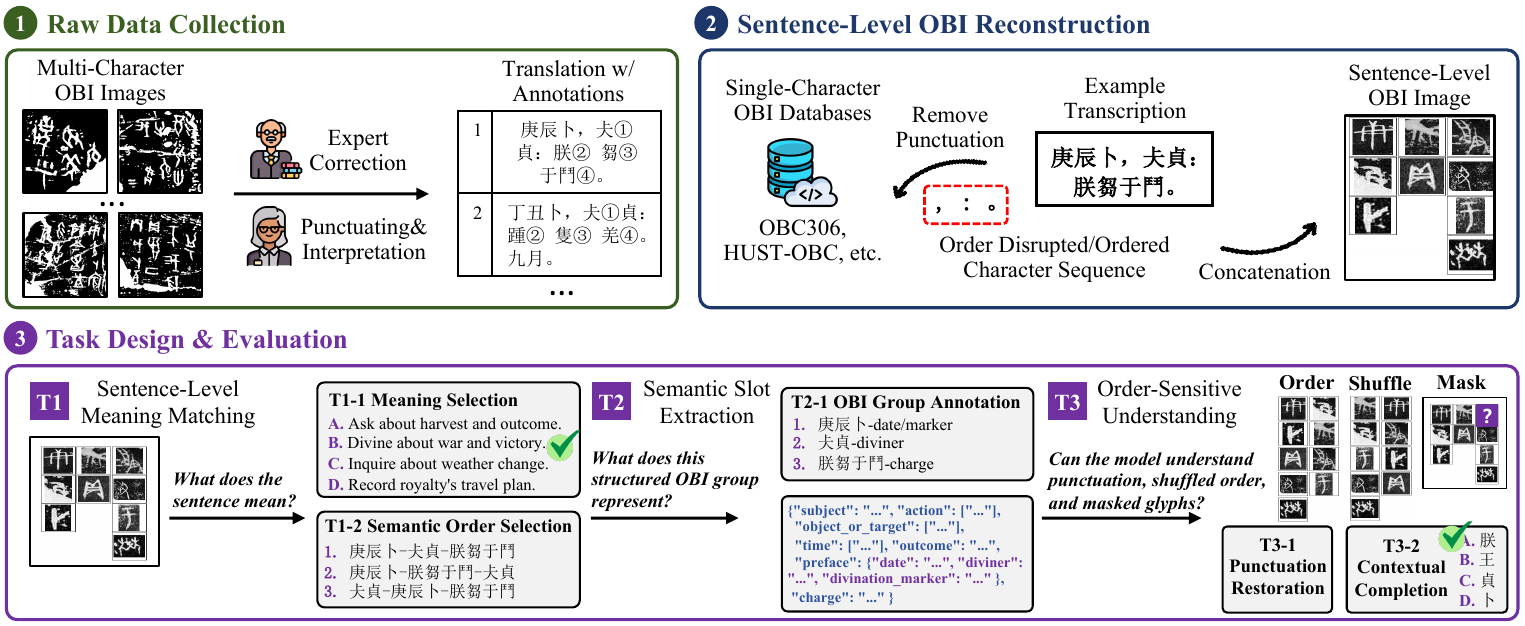}
\caption{The construction pipeline of S-OBI. We reconstruct the sentence-level OBI images based on the expert-corrected, punctuated, interpreted multi-character OBI images. Three task tiers are designed to evaluate MLLMs through semantic matching, slot extraction, and order-sensitive contextual understanding.}
\label{fig:S-OBI-overview}
\end{figure}

\section{S-OBI Benchmark}

\subsection{Task Definition}
S-OBI evaluates whether an MLLM can perceive a sentence-level OBI image and a task prompt to a semantic, structural, or contextual answer. The benchmark is designed around complete inscriptional groups rather than isolated glyphs, which therefore challenges models in combining glyph perception, character order, divination formulas, and sentence-level semantic composition.
Each evaluation instance is represented as
\begin{equation}
q_i=(X_i,p_i,a_i|t_i),
\end{equation}
where $i$ is the instance index. $X_i$ denotes the sentence-level OBI image. $p_i$ and $a_i$ denote the task prompt and the expert-corrected ground-truth answer, respectively. $t_i$ denotes the task type. Given the image-prompt pair $(X_i,p_i)$, an evaluated MLLM $M$ predicts an answer:
\begin{equation}
\hat{a}_i=M(X_i,p_i).
\end{equation}
The prediction $\hat{a}_i$ is then compared with the ground-truth answer $a_i$ according to the corresponding task type $t_i$.

\subsection{Data Curation}
\subsubsection{Source Content Collection with Expert Verification.}
High-quality original OBI rubbings and reliable translations are the foundation of S-OBI. In this work, we mainly enforce two criteria. First, the oracle bone characters on the rubbing should contain multiple characters, be sufficiently complete, and identifiable. Second, the original OBI rubbings together with their translations should be semantically complete, of a certain length, and relatively uncontroversial.
Based on this, we choose an authentic OBI book published by Liu et al. \cite{liu2005han}, and select 95 sentential OBI rubbings (Fig. \ref{visualization}(a)) as the basis for S-OBI construction.
We invited 5 OBI experts to parse, punctuate, and interpret these raw materials. 
The whole human verification process lasts over a month. Fig. \ref{fig:S-OBI-overview} illustrates the overall construction pipeline.

\subsubsection{Glyph Substitution.}
Given the scarcity of original sentence-level oracle bone inscription (OBI) materials, which are insufficient to support diverse evaluation needs, we synthesize various sentence-level OBI image sequences by replacing individual glyphs while strictly preserving the original semantic content.
After obtaining the meticulously annotated sequential OBI rubbings, we first extract individual characters from the source images. We then determine their corresponding meanings and map them to appropriate character categories using large-scale OBI datasets, such as OBC306~\cite{obc306} and HUST-OBC~\cite{hustobc}.
Within each matched category, we choose clear and reliable glyph samples as substitution candidates. These candidates are then used to generate diverse sentence-level OBI images.
The purpose of glyph substitution is not to change the semantic structure of the original inscription but to reduce accidental visual noise and degradation in the original rubbing. By replacing noisy or incomplete glyphs with clearer and more diverse corresponding samples, S-OBI preserves the original inscriptional structure while enabling a more focused evaluation of sentence-level understanding.

\subsubsection{Sentence-level OBI reconstruction.}
After obtaining candidate glyphs, we compose them according to the original character order and generate standardized sentence-level OBI images. Meanwhile, we generate controlled variants, e.g., substituted glyph sequences, shuffled character orders, and masked-glyph images, for different tasks, allowing the benchmark to evaluate different abilities while keeping the underlying inscriptional meaning consistent.

\subsection{Task Design and Difficulty Tiers}
In this section, we elaborate on the three tasks, i.e., semantic matching, semantic slot extraction, and contextual reasoning in S-OBI (Fig. \ref{fig:S-OBI-overview}).

\subsubsection{T1: Semantic matching.}
This task includes two variants, i.e., meaning selection and semantic order judgment. 
In the meaning selection task, given a sentence-level OBI image with a complete divination charge and a natural language question concerning its holistic semantic content, the model is required to select the correct answer from multiple candidates.
In the semantic order judgment task, the sentence-level OBI image is divided into several semantic units, and the model is asked to select the correct reading order from four candidate orders. This task evaluates not only whether the model can identify the reading order of the inscription, but also whether it can use the semantic meaning of the corresponding characters to recover the correct order from the visual input.

\subsubsection{T2: Semantic slot extraction.}
This task requires models to recover a structured divination frame from a sentence-level OBI image. 
Specifically, the model needs to recognize Oracle bone characters and output the contents of each slot in the structured frame, which assesses both direct identification of OBI glyphs and inference of ambiguous or problematic character types from known context. 
Compared with single-character recognition, this task evaluates whether an OBI understanding system can use known characters to infer other less recognizable characters within the same inscriptional group.

\subsubsection{T3: Contextual reasoning.}
Furthermore, to fully examine the fine-grained contextual reasoning ability, we introduce masked glyph completion and shuffled order recovery. For the first task, we randomly mask 30\% of the characters in a sentence-level OBI image and select one masked position as the target. The model is required to choose the corresponding modern Chinese character of the masked glyph from four options. For the second task, we randomly shuffle the character order in a sentence-level OBI image and ask the model to recover the correct reading order from the candidate options.
These two subtasks explicitly evaluate the understanding ability of an OBI understanding system at the sentence level and the ability to recover the correct reading order based on contextual and semantic cues.

\subsubsection{Difficulty Split.}
We adopt an intuitive division strategy, namely dividing tasks into four difficulty levels according to the number of characters in the sentence-level OBI image \cite{huang2026beyond}. 
Four subsets with different scales, i.e., samples with $2 \leq L \leq 5$, $6 \leq L \leq 8$, $9 \leq L \leq 10$, and $L \geq 11$ characters, are collected.
We discuss our observations among different difficulty levels in Sec. \ref{sec::difficulty}, where we reveal the relevance between the length of sequential OBI images and different tasks.

\section{Experiments}

\subsection{Experimental Setup}
We evaluate ten cutting-edge MLLMs, including proprietary models, GPT-5.2~\cite{gpt52}, GPT-4o~\cite{gpt4o}, Gemini 3.1 Pro~\cite{geminiapi}, and Qwen3.6-Plus~\cite{qwen36}, from OpenAI, Google, and Alibaba, as well as open-source models, DeepSeek-V4-Pro~\cite{deepseekv4}, InternVL3.5~\cite{internvl35}, Qwen3.6-35B~\cite{qwen36}, Qwen3.5-9B~\cite{qwen35}, and  Qwen2.5-VL-72B~\cite{qwen25vl} from DeepSeek, Shanghai AI Laboratory, and Alibaba.
Apart from the proprietary models and those with a parameter size greater than 72B that are deployed via API, all other models are performed using up to 8 Nvidia RTX4090 24GB. We use the default parameter settings (e.g., \texttt{temperature}, \texttt{top\_k}, and \texttt{top\_p}) of respective models.
We mainly report the overall, task-averaged, and instance-wise accuracy scores in percentage for performance comparison.

\subsection{Main Results}

\begin{table}[t]
\caption{Performance comparison on S-OBI. The upper and lower parts are closed-source and open-source MLLMs, respectively. The best and second-best results are in bold and underlined.}
\label{tab:main-results}
\centering
\scriptsize
\renewcommand{\arraystretch}{1.1}
\setlength{\tabcolsep}{3pt}
\begin{tabularx}{\textwidth}{@{}>{\raggedright\arraybackslash}p{0.36\textwidth}*{5}{>{\centering\arraybackslash}X}@{}}
\hline
Model & Overall & Avg. & T1 & T2 & T3 \\
\hline
GPT-5.2 & 24.7 & 26.8 & 55.1 & 14.3 & 11.0 \\
GPT-4o & 20.2 & 21.3 & 37.1 & 14.5 & 12.4 \\
Gemini-3.1-Pro & \underline{31.4} & \underline{32.2} & \underline{59.0} & 24.6 & 13.0 \\
Qwen3.6-Plus-thinking & \textbf{33.5} & \textbf{34.1} & \textbf{60.9} & \textbf{27.2} & \textbf{14.1} \\
\hdashline
DeepSeek-V4-Pro & 25.1 & 25.6 & 50.8 & 19.3 & 6.9 \\
Qwen3.6-35B-A3B & 26.6 & 28.7 & 56.6 & 16.2 & \underline{13.3} \\
Qwen3.5-9B & 29.7 & 29.9 & 51.5 & \underline{25.2} & 13.1 \\
Qwen2.5-VL-72B & 28.1 & 28.5 & 53.9 & 22.4 & 9.3 \\
InternVL3.5-38B & 27.3 & 28.7 & 57.2 & 18.4 & 10.6 \\
InternVL3.5-8B & 19.9 & 20.0 & 39.0 & 16.2 & 4.8 \\
\hline
\end{tabularx}
\end{table}

\noindent
\textbf{Current MLLMs are still far from reliable sentence-level OBI understanding.}
Tab. ~\ref{tab:main-results} shows that the best overall score is only 33.5\%, achieved by Qwen3.6-Plus-thinking, while Gemini-3.1-Pro ranks second at 31.4\%. All remaining models fall below 30\% overall. This ceiling indicates that S-OBI is not solved by general OCR-rich visual recognition or broad multimodal instruction following.

\noindent
\textbf{The main difficulty lies in moving from plausible meaning selection to structured and contextual interpretation.}
For the strongest model, performance drops from 60.9\% on T1 to 27.2\% on T2 and 14.1\% on T3. The same pattern appears across model families, i.e., models can often choose a semantically plausible answer, but they struggle to recover the internal divination frame and to use sentence context for punctuation or masked-glyph reasoning.

\noindent
\textbf{Open-source models are competitive, but the gap to reliable benchmark performance remains large.}
Qwen2.5-VL-72B and InternVL3.5-38B are the strongest open-source models, reaching 28.1\% and 27.3\% overall, respectively. DeepSeek-V4-Pro follows at 25.1\%. Their proximity to several closed-source systems suggests that the access regime alone does not explain performance. We speculate that the dominant bottleneck is the lack of sentence-level ancient-script grounding.

\subsection{Diagnostic Results}

\begin{table}[t]
\caption{Performance comparison on the subtasks of S-OBI.}
\label{tab:diagnostic-results}
\centering
\scriptsize
\renewcommand{\arraystretch}{1.1}
\setlength{\tabcolsep}{3pt}
\begin{tabularx}{\textwidth}{@{}>{\raggedright\arraybackslash}p{0.34\textwidth}*{6}{>{\centering\arraybackslash}X}@{}}
\hline
Model & Avg. & Meaning & Order & Slots & Punc. & Mask \\
\hline
GPT-5.2 & 29.3 & 27.4 & 82.8 & 14.3 & 10.5 & 11.7 \\
GPT-4o & 23.3 & 11.6 & 62.7 & 14.5 & 7.7 & \textbf{20.0} \\
Gemini-3.1-Pro & \underline{33.5} & \underline{32.6} & 85.3 & 24.6 & \textbf{14.9} & 10.0 \\
Qwen3.6-Plus-thinking & \textbf{35.6} & \underline{32.6} & \textbf{89.2} & \textbf{27.2} & \underline{12.5} & \underline{16.7} \\
\hdashline
DeepSeek-V4-Pro & 27.1 & 13.7 & \underline{87.8} & 19.3 & 4.9 & 10.0 \\
Qwen3.6-35B-A3B & 31.3 & 26.3 & 86.9 & 16.2 & 12.3 & 15.0 \\
Qwen3.5-9B & 31.2 & 25.3 & 77.8 & \underline{25.2} & 10.8 & \underline{16.7} \\
Qwen2.5-VL-72B & 29.7 & 25.3 & 82.5 & 22.4 & 9.9 & 8.3 \\
InternVL3.5-38B & 30.6 & \textbf{33.7} & 80.7 & 18.4 & 12.0 & 8.3 \\
InternVL3.5-8B & 20.8 & 17.9 & 60.1 & 16.2 & 4.7 & 5.0 \\
\hdashline
Task-wise Avg.&29.24 & 24.64 &79.58 & 19.83 &10.02& 12.17 \\
\hline
\end{tabularx} 
\end{table}

\begin{figure}[!t]
\centering
\includegraphics[width=.6\textwidth]{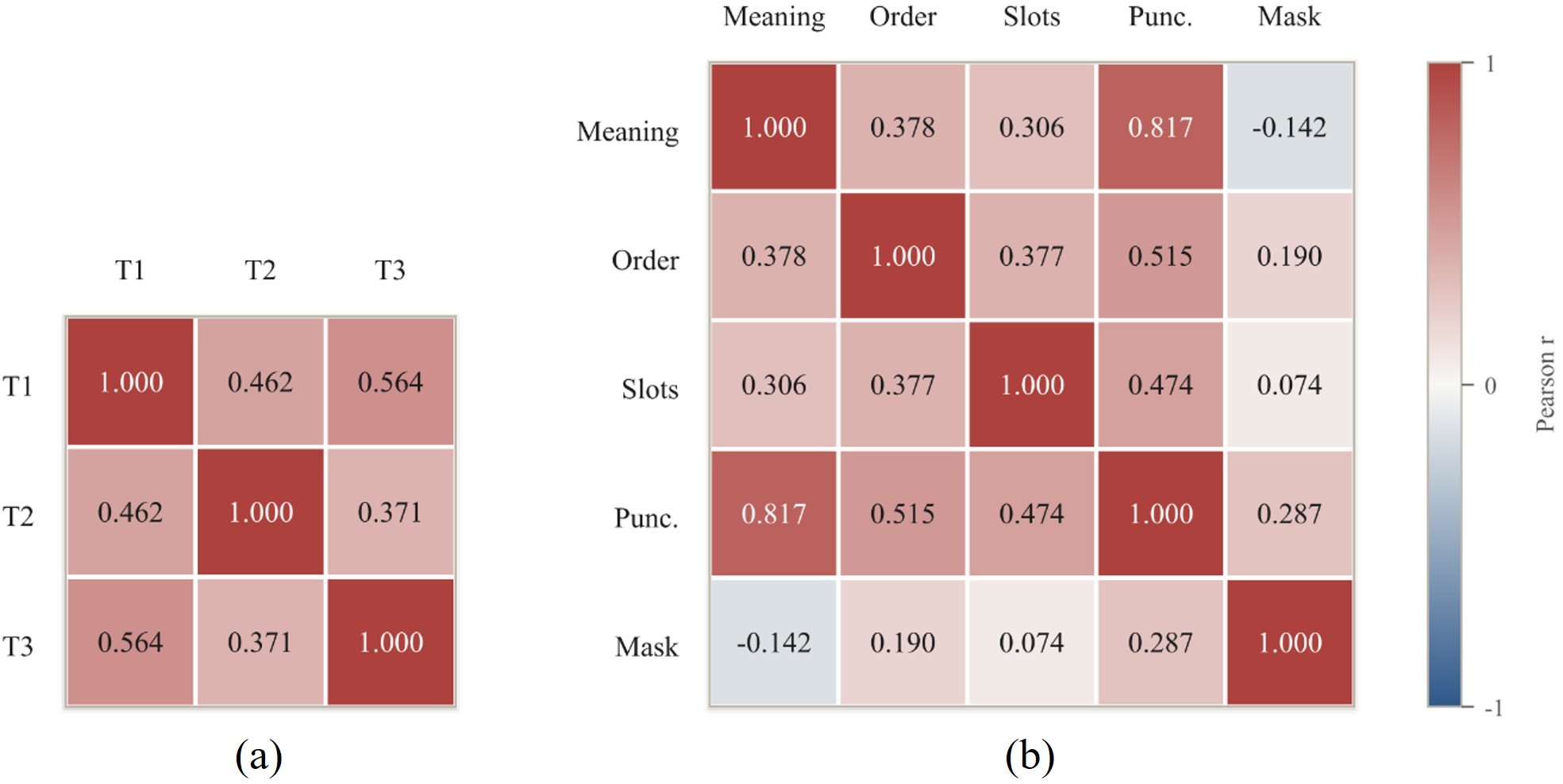}
\caption{Pearson's correlation between different tasks.}
\label{fig:correlation}
\end{figure}

\noindent
\textbf{Models capture coarse order more easily than sentence-level meaning.}
Tab. ~\ref{tab:diagnostic-results} reports an average score of 79.6\% for semantic-unit order selection, but only 22.6\% for meaning matching. This contrast is visible even in strong runs. For example, Qwen3.6-Plus-thinking reaches 89.2\% on order selection but 32.6\% on meaning matching, while DeepSeek-V4-Pro reaches 87.8\% on order selection but only 13.7\% on meaning matching. This suggests that visible sequence regularities are easier to exploit than inscription-level semantic interpretation.

\noindent
\textbf{Structured slot extraction remains a central bottleneck rather than a formatting issue.}
The average slot score is nearly 20.0\%, and the best score is 27.2\%. Although several models produce syntactically valid structured outputs, all models obtain 0\% complete slot-frame exact match. This gap shows that models often miss at least one essential component of the divination structure, such as the subject, action, target, time, outcome, preface, or charge.

\noindent
\textbf{Order-sensitive contextual recovery is especially weak.}
Punctuation restoration remains below 15\% for every model, with Gemini-3.1-Pro achieving the best score at 14.9\%. Besides, masked glyph completion is also fragile. The best result is below the nominal 25\% chance level of a four-choice task. These failures indicate that current MLLMs can not yet use visible context and inscriptional order robustly when the local glyph evidence is incomplete.

\noindent
\textbf{Correlation analysis.}
Fig. \ref{fig:correlation} shows Pearson's correlation between different tasks. We observe moderate interdependence ($r \in [0.371, 0.564]$) in macro-level correlations (Fig. \ref{fig:correlation}(a)), demonstrating the rationality of our task design and the reduction of redundancy. Granular subtask analysis (Fig. \ref{fig:correlation}(b)) reveals a strong semantic-syntactic coupling ($r = 0.817$) between "Meaning" and "Punc.", indicating that LMMs heavily depend on overall semantic decoding to segment unpunctuated ancient scripts. 
Structural reasoning tasks, such as "Order" and "Slots", align moderately, suggesting shared downstream pathways for grammatical arrangement. 
Conversely, the "Mask" task acts as an orthogonal outlier (e.g., $r = -0.142$ with Meaning), proving that masked character restoration prioritizes local visual pattern matching over broad semantic logic. 
Consequently, robust evaluation frameworks must strategically pair these correlated semantic tasks with orthogonal visual tasks to thoroughly assess both high-level text decipherment and low-level visual robustness.

\begin{table}[t]
\caption{Difficulty-conditioned results. $\Delta$ denotes `Very Hard' minus `Easy'.}
\label{tab:difficulty-results}
\centering
\scriptsize
\renewcommand{\arraystretch}{1.1}
\setlength{\tabcolsep}{3pt}
\begin{tabularx}{\textwidth}{@{}>{\centering\arraybackslash}p{0.12\textwidth}>{\raggedright\arraybackslash}p{0.24\textwidth}*{5}{>{\centering\arraybackslash}X}@{}}
\hline
Task & Profile & Easy & Medium & Hard & Very hard & $\Delta$ \\
\hline
T1 & Mean & 41.3 & 45.2 & 50.5 & 50.1 & +8.8 \\
T1 & Qwen3.6-Plus-thinking & 58.9 & 56.9 & 65.2 & 62.3 & +3.4 \\
\hdashline
T2 & Mean & 18.4 & 16.7 & 14.2 & 13.8 & -4.6 \\
T2 & Qwen3.6-Plus-thinking & 36.8 & 28.7 & 19.1 & 20.4 & -16.4 \\
\hline
\end{tabularx}
\end{table}

\begin{figure}[!t]
\centering
\includegraphics[width=1\textwidth]{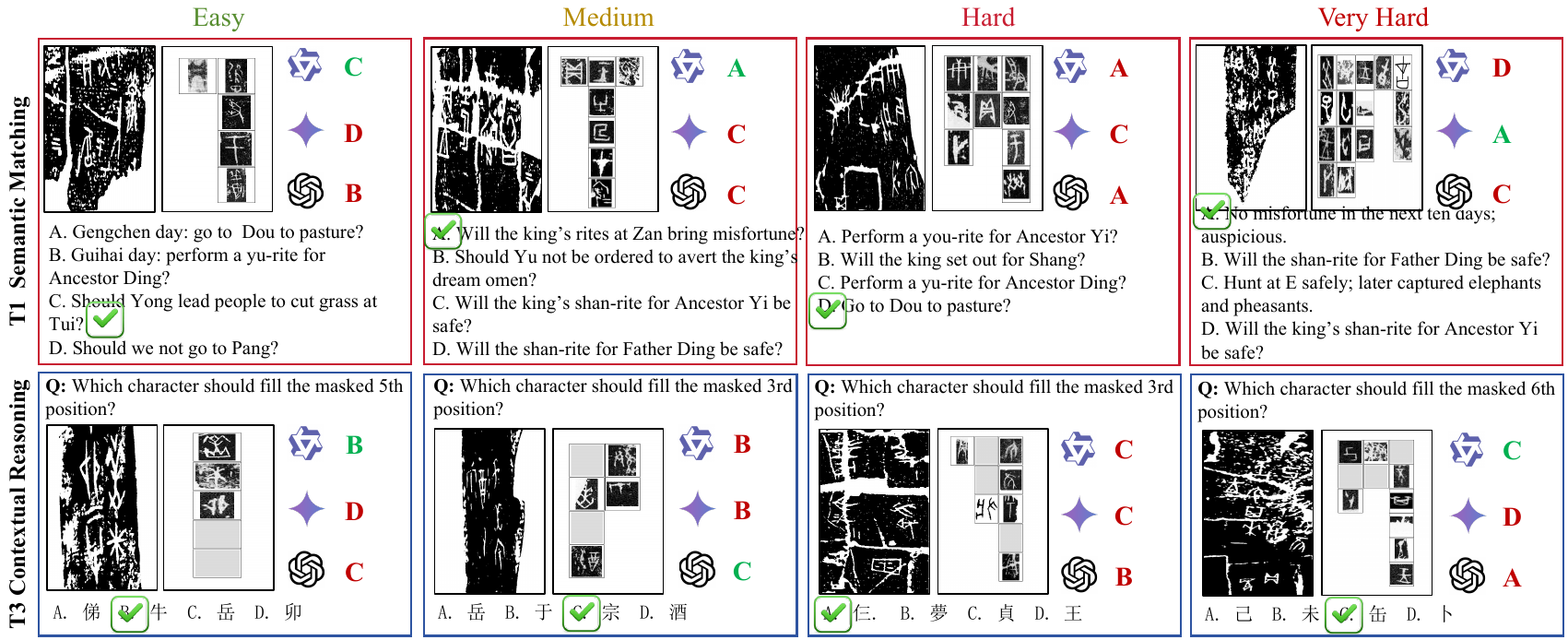}
\caption{Case studies on semantic matching and contextual reasoning tasks across different difficulties.}
\label{fig:case}
\end{figure}

\subsection{Difficulty Analysis}
\label{sec::difficulty}
Tab. ~\ref{tab:difficulty-results} shows that character length affects different task families in opposite ways. T1 does not decrease with length, suggesting that longer inscriptions may expose more formulaic ordering cues. By contrast, T2 degrades as length increases. Specifically, the average score drops by 4.6 points from easy to very hard, and Qwen3.6-Plus-thinking drops by 16.4 points. This supports the interpretation that length is a pressure test for structured role assignment rather than a simple visual-complexity factor.
Fig. \ref{fig:case} provides a qualitative comparison between Qwen3.6-plus-thinking, Gemini 3.1 Pro, and GPT-4o.

\section{Limitations}
Although we have conducted extensive exploration on S-OBI and obtained important observations, S-OBI has several limitations. First, the current benchmark is compact, with 95 base inscriptions, and is therefore intended as a diagnostic benchmark rather than a large-scale leaderboard. 
Second, the sentence-level images are reconstructed from aligned glyph images, which simplifies some material properties of original oracle bones. 
Third, multiple-choice subtasks may still contain answer-choice shortcuts, especially for formulaic order selection.

\section{Conclusion}
In this paper, we introduce S-OBI, a novel benchmark for evaluating MLLMs on sentence-level oracle bone inscription image understanding. Unlike single-character OBI resources, S-OBI targets evaluating the inscriptional-group understanding through global meaning matching, semantic slot extraction, punctuation restoration, and masked glyph completion. 
Experiments on ten mainstream MLLMs show that current systems remain unreliable in sentence-level OBI understanding, with the best overall performance being 33.5\%.
Subtask analysis further shows that MLLMs perform much better on semantic-unit order selection than on full meaning matching, suggesting reliance on coarse formulaic regularities rather than robust inscriptional reasoning. 
These findings position sentence-level ancient-script understanding as a distinct and challenging multimodal evaluation problem for language models.

\end{document}